\documentclass[a4paper,fleqn]{cas-sc}

\usepackage[authoryear,longnamesfirst]{natbib}

\def\tsc#1{\csdef{#1}{\textsc{\lowercase{#1}}\xspace}}
\tsc{WGM}
\tsc{QE}
\tsc{EP}
\tsc{PMS}
\tsc{BEC}
\tsc{DE}
\begin{document}
\let\WriteBookmarks\relax
\def\floatpagepagefraction{1}
\def\textpagefraction{.001}
\shorttitle{Deep Multi-Modal Sensor Fusion using Fusion Weight Regularization and Target Learning}
\shortauthors{MS Shim et~al.}
%\begin{frontmatter}

\title [mode = title]{Robust Deep Multi-Modal Sensor Fusion using Fusion Weight Regularization and Target Learning}                      
% \tnotemark[1,2]

%\tnotetext[1]{This document is the results of the research project funded by the National Science Foundation.}

%\tnotetext[2]{The second title footnote which is a longer text matter to fill through the whole text width and overflow into another line in the footnotes area of the first page.}

\author[1]{Myung Seok Shim}
\ead{mrshim1101@tamu.edu}
\address[1]{Department of Electrical and Computer Engineering, Texas A\&M University, College Station,
TX, 77843, USA}

\author[2]{Chenye Zhao}
\ead{czhao43@uic.edu}
\address[2]{Electrical and Computer Engineering, University of Illinois at Chicago,
851 S Morgan St, Chicago, IL 60607, USA}

\author[1]{Yang Li}

\ead{li13157@tamu.edu}

\author[3]{Xuchong Zhang}
\fnmark[1]
\ead{zhangxc@stu.xjtu.edu.cn}
\address[3]{School of Electronic and Information Engineering Xi'an Jiaotong University,
Xianning West Road, Xi'an, Shaanxi, 710049, P.R. China}

\author[4]{Wenrui Zhang}
\ead{wenruizhang@ucsb.edu}

\author[4]{Peng Li}
\cormark[1]
\ead{lip@ucsb.edu}

\address[4]{Department of Electrical and Computer Engineering, UC Santa Barbara, Santa Barbara, CA 93106, USA}

\cortext[cor1]{Corresponding author}
\fntext[fn1]{The work was performed during his visit to Texas A\&M University.}

\begin{abstract}
Sensor fusion has wide applications in many domains including health care and autonomous systems.  While the advent of deep learning has enabled promising multi-modal fusion of high-level features and end-to-end sensor fusion solutions, existing deep learning based sensor fusion techniques including deep gating architectures are not always resilient, leading to the issue of fusion weight inconsistency, sensory noise vulnerability, and feature importance interpretability. We propose deep multi-modal sensor fusion architectures with enhanced robustness particularly under the presence of sensor failures. At the core of our gating architectures are fusion weight regularization and fusion target learning operating on auxiliary unimodal sensing networks appended to the main fusion model. The proposed regularized gating architectures outperform the existing deep learning architectures with and without gating under both clean and corrupted sensory inputs resulted from sensor failures. The demonstrated improvements are particularly pronounced when one or more multiple sensory modalities are corrupted.
\end{abstract}

\iffalse
\begin{graphicalabstract}
\includegraphics{figs/grabs.pdf}
\end{graphicalabstract}

\begin{highlights}
\item Research highlights item 1
\item Research highlights item 2
\item Research highlights item 3
\end{highlights}
\fi


\maketitle

\section{Introduction}
Sensor fusion is to combine multiple sensory inputs to create accurate ground truth in complex systems. With advent of deep learning, the sensor fusion algorithm is being improved and applied on applications such as autonomous systems equipped with sensing modalities \citet{ramachandram2017deep}. For instance, inputs from multiple cameras and LIDARs are fused for autonomous driving with deep learning technologies\citet{ku2017joint,chen2017multi,wei2018LIDAR}. In addition to the application, smart watches are attached with many sensors: GPS, accelerometer, gyroscope, ECG, blood oxygen sensors, etc. The deep sensor fusion algorithm in these devices are applied for checking users' health, recognizing posture and predicting heart attack. \citet{dehzangi2017imu, mauldin2018smartfall, ali2020smart}.

For creating meaningful outputs from the multiple sensory inputs, elaborate neural network architecture is needed for the sensor fusion. Since the number of sensory inputs is increasing, the way how to fuse the inputs is important and studied \citet{patel2017sensor,mees2016choosing, 7837868}. Due to many number of sensors utilized for the applications above, discovering important features and other redundant features is essential for better target prediction. The prediction performance may be boosted by weighting the important features.

\begin{figure}[htbp]
\centering
\includegraphics[width=0.7\linewidth]{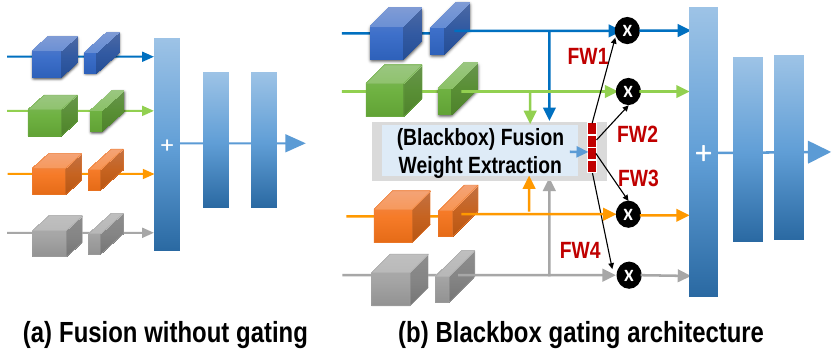}
\caption{Two DNN sensor fusion architectures.}
\label{fig:Architectures}

\end{figure}

Nowadays, convolutional neural network based (CNN) approaches are proposed to the sensor fusion. For instance, \citet{chen2017multi} presents a CNN based algorithm in autonomous driving and compares three fusion schemes: 
early, late, and deep fusion. 
However, this work does not consider sensor failures and provides no deep insight on computational principles of sensor fusion.
The typical late fusion DNN architecture is shown in Fig.~\ref{fig:Architectures} (a) where each modality is first processed, e.g. by a number of convolutional/pooling/activation layers. And then all extracted feature maps are fused together, e.g. by element-wise mean at the layer marked by ``+'', and processed further to make the final decision.    

A key objective and challenge in sensor fusion is attaining \emph{resilience}, i.e., the  fusion task shall be conducted robustly not only under clean sensory inputs but also in the presence of  sensor failures. % Towards this end, gating fusion architectures are of particular interest. 
\citet{zhang2019robust}, \citet{kendall2018multi}, and \citet{gao2019multi} display CNN based object tracking with input weighting (gating) fusion scheme and auxiliary losses. By weighting more important input channel with fusion weights (gating factors) which are processed with a sigmoid function, the weighted important inputs have more effect on outputs and the classification accuracy is further improved.

\citet{arevalo2017gated} proposes a gated multi-modal unit (GMU) for data fusion with two or more modalities. The GMU identifies how modalities affects the network in activation function level by using multiplicative gates. However, this proposed algorithm could not show the feature importance through unit activation and the interpretability of the learned features is not examined in this work. 

As another example of gating architectures, \citet{patel2017sensor} proposes a gated convolutional neural network called \emph{NetGated} architecture, where two fusion weights which are extracted from the camera and LIDAR inputs are employed to multiply the corresponding pre-processed camera/LIDAR features, respectively, in order to compute a weighted sum of the two. The weighted sum passes through fully connected layers to produce the final robot steering command. This gating architecture is illustrated for the case of four sensory modalities in Fig.~\ref{fig:Architectures}(b), where four scalar fusion weights (FW1 to FW4) are extracted by fusing the pre-processed feature maps of the four modalities. % The gating architecture  in \cite{kim2018robust} is in the same spirit of the NetGated architecture with the main difference being fusion weight maps are extracted. 

%A weight map is extracted for each sensory modality from the concatenated  multi-sensory feature maps produced at each convolutional pre-processing step. Each weight map is multiplied with the feature maps of the same modality. These weighted feature maps are concatenated and further processed to produce a fused feature map.

Comparing with a non-gating architecture like the one in Fig.~\ref{fig:Architectures}(a), gating architectures (e.g. Fig.~\ref{fig:Architectures}(b)) are positioned to provide a more structured fusion solution. Interference from a highly noisy or failing sensor can be completely gated off simply when the corresponding extracted fusion weight is zero, a desirable property  hard to be implemented in a non-gating architecture.

% Main contribution

In this paper, we propose a new gating architecture and its variants with significantly improved performance and resilience under both clean sensory inputs and sensor failures. Our main contributions are: \textbf{1)} propose a new auxiliary-model regulated gating architecture, called \emph{ARGate}, to robustly learn gating fusion weights of different modalities using auxiliary unisensory processing paths during training; \textbf{2)} As part of the ARGate architecture, propose two regularization techniques, namely, \emph{fusion weight regularization with auxiliary losses}, and \emph{monotonic fusion target learning} to regularize fusion weights with corresponding auxiliary losses and target for consistent fusion weights in the training process and significantly improve the performance and robustness; and \textbf{3)} shed light on the fusion and regularization mechanisms that are responsible for the observed performance improvements.

We perform comprehensive evaluation of the proposed ARGate architectures while comparing with a baseline non-gating CNN architecture and the NetGated architecture~\cite{patel2017sensor} utilizing the Human Activity Recognition (HAR) dataset~\cite{anguita2013public}, driver identification dataset~\cite{kwak2016know}, and KITTI dataset~\cite{Geiger2012CVPR} under various sensory input conditions. It is demonstrated that the proposed architectures consistently outperform the baseline and NetGated architectures, and  improve  classification accuracy by up to $8.49\%$ and $13.39\%$ over the baseline and NetGated architectures, respectively. For the KITTI dataset, the proposed architecture outperforms a reference architecture \cite{ku2017joint} by up to 4.81\%.

% Main problem : fusion weight inconsistency, sensory noise vulnerability, and feature importance interpretability
\section{Limitations of the existing gating architectures}
\subsection{Fusion weight inconsistency}

Gating architectures such as NetGated extract a fusion weight (map) for each sensory input \citet{patel2017sensor, mees2016choosing, kim2018robust}. Ideally,
the fusion weights shall reflect the integrity and importance of the corresponding modalities w.r.t. the classification task. It is desirable to gate off a corrupted modality via a small or zero-valued fusion weight. However, the end-to-end black-box nature of these  architectures can lead to \emph{fusion weights inconsistency}. Fusion weights which are extracted by the NetGated \citet{patel2017sensor}  tend to be unstable and fail to consistently reflect the importance of  modalities, leading to observed poor performance. There are cases where the Netgated architecture even underperforms non-gating architectures. Furthermore, gating architectures with the auxiliary losses \citet{kendall2018multi, zhang2019robust} which are handled in each auxiliary input channel regularize a main CNN model. Though this regularization scheme improves overall performance, how it regularizes the fusion weight is not examined. Because this fusion weight regularization may resolve fusion weight inconsistency issue with target fusion weight learning, the classification accuracy might be further improved.

\subsection{Sensory noise vulnerability}
Similar to fusion weights inconsistency, gating architectures are required to handle \emph{sensory noise vulnerability}. In real world, some sensory inputs are significantly important for output prediction, while the other sensors are not due to the noise and the environmental difference. For example, an autonomous vehicle is equipped with a radar, a Lidar, and cameras which contribute to the overall performance of driving and drivers' safety. Even if one of the sensors is malfunctioning, the autonomous driving performance should be stable with other two sensors by gating off the broken sensor. However, the above-mentioned sensor fusion architectures are well not considered for the noise vulnerability. 

\subsection{feature importance interpretability}
Lastly, the core of the gating architectures is \emph{feature importance interpretability} via fusion weights. Due to the black box nature of neural networks, the importance of feature was un-interpretable. The Netgated architecture is designed for weighting more important features, but not well performed because of the fusion weight inconsistency. Furthermore, since the fusion weights are not normalized, the fusion weight interpretability is not demonstrated. Through the proposed ARGate architecture in the next section, these three problems are resolved with the proposed regularization technique and monotonic fusion target learning scheme.

\section{Overview of the ARGate Architectures}
\begin{figure}[h]
\centering
\includegraphics[width=0.5\linewidth]{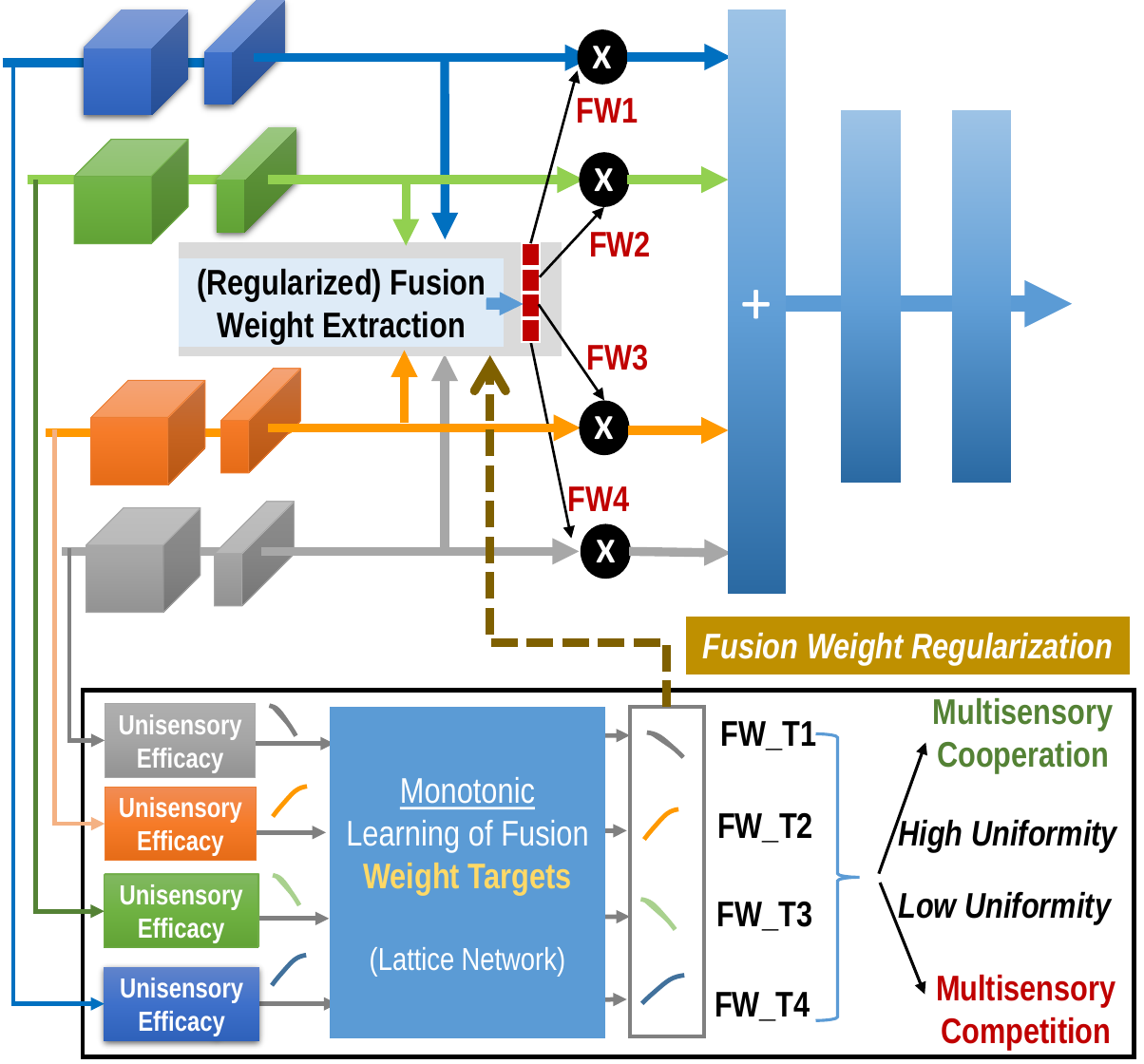}
\caption{ARGate architecture with fusion weight regularization and monotonic fusion weight target learning. The bottom box offers fusion weight regulation for training the main model (upper portion) and is removed for inference.}
\label{fig:overview}
\end{figure}

The ARGate architecture with the two most essential regularization techniques, i.e.  \emph{fusion weight regularization with auxiliary losses} and \emph{monotonic fusion weight target learning}, is depicted in Fig.~\ref{fig:overview}.

Multisensory processing has been under extensive study in neuroscience and recent research has shed increasing light on the computational mechanisms underlying the  multi-sensory processing  of superior colliculus (SC) neurons in the midbrain
\citet{Yu:JNEUROSCI:2019,Stein:NatureNeuro:2014}, providing a relevant reference for the proposed ARGate architectures. Animals must promptly select the most pertinent modalities while suppressing spurious noise from other senses for their survival, where \emph{multisensory competition} plays an essential role.  This is analogous to the targeted robust sensor fusion problem under catastrophic sensor failures. 
On the other hand, mature SC neurons are able to enhance their performance via \emph{multisensory cooperation} where congruent stimuli across modalities are integrated. This corresponds to performance benefits in the targeted  sensor fusion  brought by fusing co-variant and complementary sensory inputs. 

As such, multisensory competition and cooperation may be considered as two intertwined operating modes, transitions between which may be orchestrated based on sensory conditions: operate in competition when inputs are non-congruent (e.g. due to strong sensory noise or catastrophic failures) to ensure robustness while switching to cooperation to maximize performance with clean/covariant inputs. 

Loosely speaking, the proposed architecture in Fig.~\ref{fig:overview} (a detailed full implementation is in 
Fig.~\ref{fig:lattice}) bears high-level resemblance to  mode transitions in SC neurons %at different developmental stages 
suggested by  the electrophysiological recording and computational studies  \citet{Yu:JNEUROSCI:2019}. However, it takes a rather different strategy to balance between multisensory competition and cooperation in the end-to-end deep learning architecture during training.   The bottom block of  Fig.~\ref{fig:overview} outputs the fusion weight targets (FW\_T1 to FW\_T4), which are used to regularize the fusion weight extraction block such that each fusion weight  FW\_i is constrained to be near the corresponding target FW\_Ti. This acts as a solution to the fusion weight inconsistency problem of the typical gating architectures. 

The key idea here is to transparently balance between multisensory competition and cooperation  by detecting the efficacy (i.e. conditioning or importance) of each unisensory input for the same end prediction task. The efficacy is evaluated by the training loss of a ``unisensory efficacy" block, which is essentially a unisensory model for the same task with the corresponding  modality being the sole input.  Intuitively, a high unisensory efficacy  shall be mapped to a high fusion weight target value. To allow for a flexible end-to-end architecture, the efficacies are mapped to the fusion weight targets (FW\_T1 to FW\_T4) by a trainable monotonic lattice network ensuring the monotonic relationship between the two. Since  fusion weight targets sum up to 1.0, low uniformity among  fusion weight targets immediately leads to multisensory competition where the inputs with smaller fusion weight targets tend to be depressed.
High uniformity in the fusion weight targets would give rise to multisensory cooperation where  modalities with similar fusion target values are integrated in a balanced way. The design of the ARGate architectures is detailed as follows.

\section{The Proposed ARGate Architecture}
\subsection{Basic structure of ARGate}

\iffalse
\begin{figure}[htbp]
\centering
  \includegraphics[width=0.8\linewidth]{basic_argate_new.pdf}
  \caption{The basic ARGate structure}
  \label{fig:auxiliary}
\end{figure}
\fi 

As a first step, we present a primitive ARGate architecture with weight sharing (ARGate-WS), which is enclosed in a more complete version of the ARGate architecture called ARGate+ in Fig.~\ref{fig:complete_argate}. The ARGate+ architecture will be built upon ARGate-WS and explained in the next subsection.

For ease of discussion, we assume that there exist two sensory inputs as in Fig.~\ref{fig:complete_argate}. 
To assist the training of the main model, which is employed for inference, an auxiliary (aux) model is included. The main model is architecturally similar  to the NetGated architecture. When splitting the output of the``FC-con" layer into the fusion weights, we further introduce sigmoid and a softmax function to normalize the fusion weights between [0,1], so that the importance of input features are expressed with the magnitude of the fusion weights. 

In general, the aux model consists of multiple independent auxiliary paths, one for each modality without fusion. 
The weight parameters in the convolutional layers and early FC layers are shared between the corresponding modalities across the main and aux models.  The total training loss is a weighted sum of the losses of the main model and all auxiliary paths. The adopted weight sharing (WS) is illustrated by the dashed purple box in Fig.~\ref{fig:complete_argate}. 
Weight sharing is commonly used in the literature \citet{chen2017multi, zhang2019robust}. Here, it acts as a way of regularizing the main model.

Based on this primitive ARGate-WS architecture only with weight sharing, we explore the more complete ARGate+ architecture with two fusion weight regularization techniques next.

\subsection{Fusion Weight Regularization with Auxiliary Losses: ARGate+ Architecture}
When one or more sensory channels are completely corrupted, weight sharing in ARGate-WS fails to properly regularize the corresponding convolutional/FC layers for the corrupted input channels in the main model. This is because that these auxiliary paths can no longer be trained to deliver a good performance based on the single corrupted modality.  

\begin{figure}[h]
\centering
\includegraphics[width=0.5\linewidth]{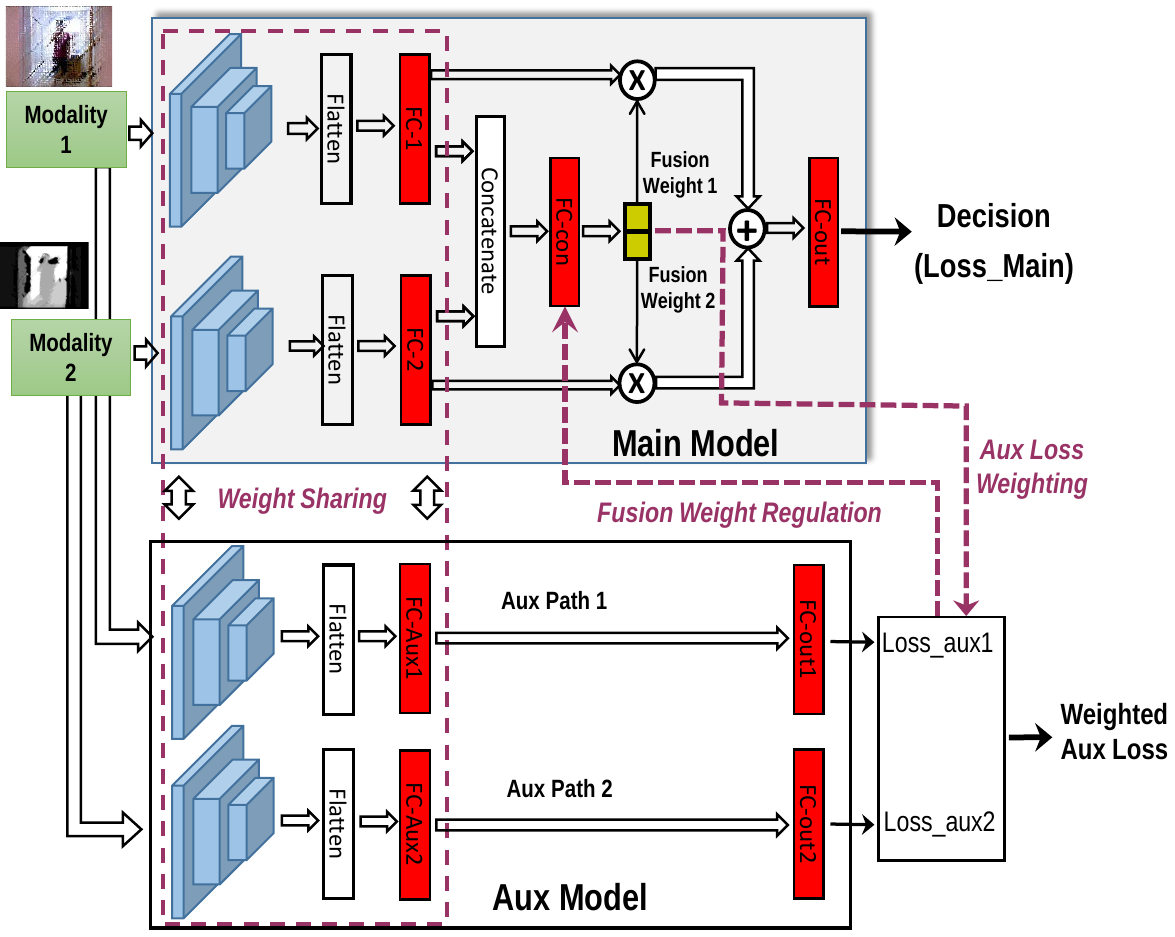}
\caption{The proposed ARGate+ architecture}
\label{fig:complete_argate}
\end{figure} 

Our key observation is that the losses of different auxiliary paths reflect the relevance of these modalities w.r.t. to the classification task, and hence, can be used for \emph{fusion weight regularization} (FWR), which is depicted using a purple dashed line pointing from the aux model to the main model in Fig.~\ref{fig:complete_argate}. When a particular sensory modality is corrupted under an input example, the high loss of its auxiliary path will constrain the training of the ``FC-con" layer in the main model to produce a low fusion weight for that modality. % We realize fusion weight regulation (FWR) by adding an additional term to the training loss:  

Since the loss of each auxiliary path is included in the total training loss,  the large loss of any corrupted sensory input can be dominant, potentially lowering performance. To resolve this problem, we use the extracted fusion weights from the main model as the corresponding weights of the auxiliary losses in the total training loss.  We call this scheme auxiliary loss weighting (ALW) as shown by the dashed purple line pointing from the main model to the aux model in Fig.~\ref{fig:complete_argate}. This leads to a more complete ARGate+ architecture  with a new loss function:

\begin{eqnarray}
\label{three_term_loss}
    Loss = & & {Loss_{main}} +  \alpha \cdot \sum_{k=1}^K {w}_{fusion}^ k {Loss_{aux}^k} + \beta \cdot \sum_{k=1}^K \left({w}_{fusion}^ k- \widehat{e^{{-{{Loss}_{aux}^k}}^2}}\right)^2
\end{eqnarray}
where the auxiliary path loss ${Loss_{aux}^k}$ is weighted by the fusion weight  ${w}_{fusion}^ k$, hence the dominance of any large auxiliary path loss resulted from sensor failures is diminished.

Each normalized auxiliary-path loss $\widehat{e^{{-{{Loss}_{aux}^k}}^2}}$ is employed to regularize ${w}_{fusion}^ k$. Here, $\widehat{e^{{-{{Loss}_{aux}^k}}^2}}$ is obtained by plugging ${Loss_{aux}^k}$ into the exponential function and then normalizing $e^{{-{{Loss}_{aux}^k}}^2}$ using sigmoid and then softmax normalization so that $\widehat{e^{{-{{Loss}_{aux}^k}}^2}}$ is between [0,1]. 

With above-mentioned explanation, the last term of the right hand side of (\ref{three_term_loss}) is directly used as fusion weight targets for regularizing the fusion weights of the main model. The ARGate+ architecture integrates weight sharing (WS), fusion weight regularization with auxiliary losses (FWR), and auxiliary loss weighting (ALW).  Next, we discuss the other key proposed technique, namely,  monotonic fusion target learning.   
% The architecture with both weight sharing (WS) and fusion weight regularization (FWR) is dubbed ARGate-WS-FWR.

\iffalse

\begin{eqnarray}\label{eqn_fwr}
\label{three_term_loss_without_Fusion_weights}
    Loss = & &  {Loss_{main}} + \alpha \cdot \sum_{k=1}^K {Loss_{aux}^k} \\ \nonumber
    & & + \beta \cdot \sum_{k=1}^K \left({w}_{fusion}^ k- \widehat{e^{{-{{Loss}_{aux}^k}}^2}}\right)^2,

\end{eqnarray}
\fi
%norm(e^{{-{Loss}_{aux}^k}^2})
% where $\beta$ is another user-defined weighting factor,  and ${w}_{fusion}^ k$ is the fusion weight for the $k$-th modality outputted by the ``FC-con" layer in the main model. 

%While ARGate-WS and ARGate-WS-FWR both  improve performance, the full ARGate architecture (ARGate+) with both fusion weight regularization and auxiliary loss weighting consistently produces the best performance, as will be demonstrated.   

\subsection{Monotonic Fusion Target Learning}
Regularizing fusion weight through auxiliary losses is a key step towards improving the overall performance. Notice that  the last term in (\ref{three_term_loss}) implements fusion weight regularization in which the exponential of each normalized auxiliary path is called the  fusion weight target for the corresponding fusion weight. While this specific form of fusion weight targets pushes the training of the network towards producing a low fusion weight value for modalities with a large auxiliary path loss, the optimal mapping from the auxiliary path loss to the corresponding fusion weight target is not known \emph{a priori}.   Our key idea is to optimize this mapping end-to-end as part of the overall network architecture based on the available training data.

Since the fusion weight target of an auxiliary path shall be a \textit{monotonically} non-increasing function of the corresponding auxiliary path loss, this motivates us to introduce a dedicated small network to learn the mapping from the set of auxiliary path losses $ \vec{L}oss_{aux} = [ Loss^1_{aux},  Loss^2_{aux}, \cdots, Loss^K_{aux}]^T$ to each fusion weight target $w^k_{fusion,t}$: $w^k_{fusion,t} = f_k(\vec{L}oss_{aux})$. Corresponding, the loss is modified from (\ref{three_term_loss}) to:

\begin{eqnarray}
\label{three_term_loss_lattice}
    Loss = & & {Loss_{main}} +  \alpha \cdot \sum_{k=1}^K {w}_{fusion}^ k {Loss_{aux}^k}  + \beta \cdot \sum_{k=1}^K \left({w}_{fusion}^ k- f_k(\vec{L}oss_{aux})\right)^2.
\end{eqnarray}

\begin{figure}[h]
\centering
\includegraphics[width=0.5\linewidth]{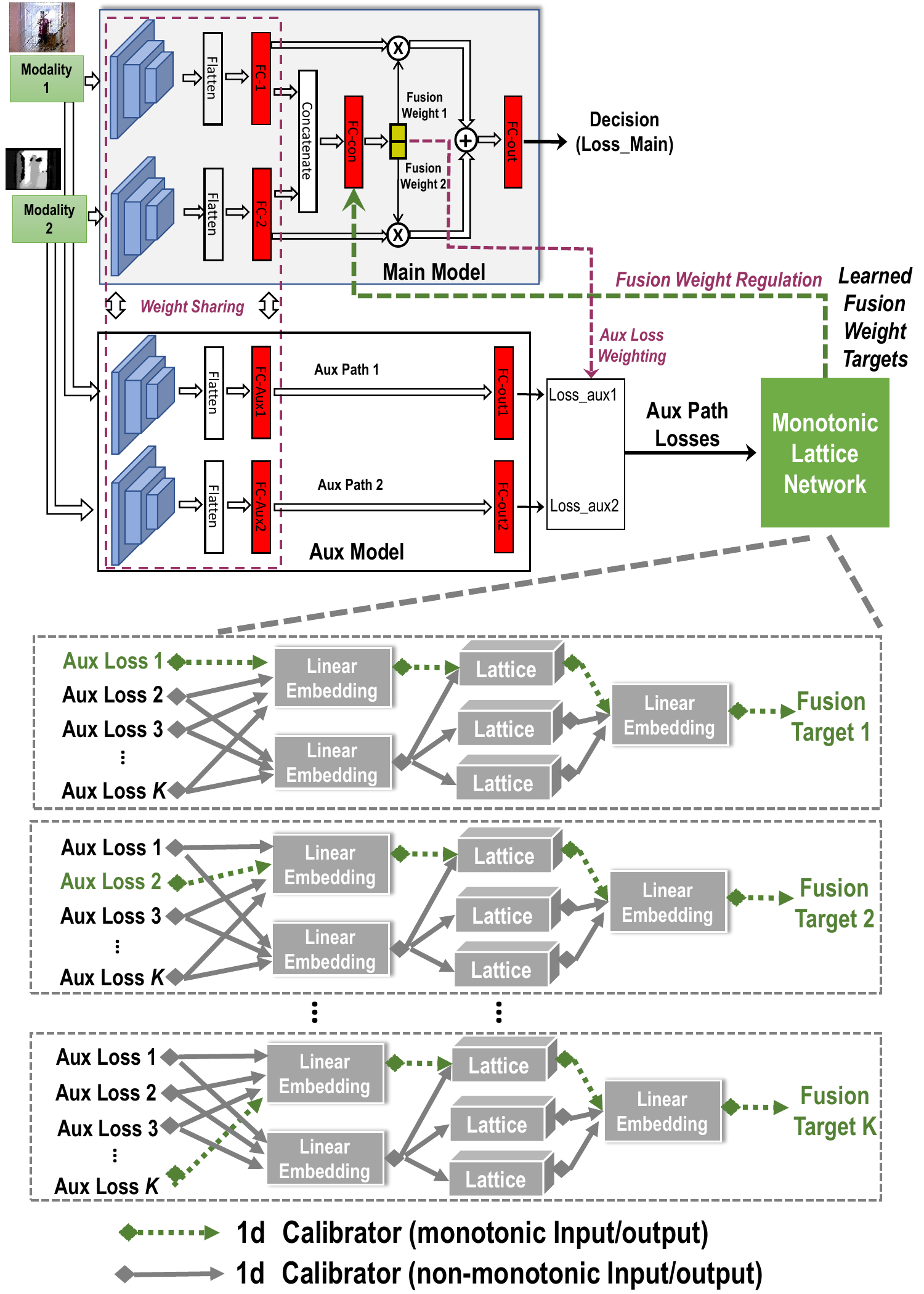}

\caption{The proposed ARGate-L architecture with end-to-end monotonic learning of fusion targets using a lattice network.}

\label{fig:lattice}
\end{figure}

The mappings from the auxiliary losses to the fusion weight targets may be learned with great flexibility by a generic feed-forward multi-layer  neural network. However, this approach is not robust  due to lack of regularization. We propose to embed a regularized deep lattice network (DLN) \cite{lattice} to the ARGate+ architecture for more robust learning of fusion targets. As illustrated in the lower part of Fig. \ref{fig:lattice}, a lattice network can learn input-to-output mappings while guaranteeing the user-specified full or partial monotonicity between the set of the inputs and outputs.

\iffalse
\begin{figure}[htbp]
	\centering
	\includegraphics[width=0.5\linewidth]{lattice_basic.pdf}

	\caption{Learning of monotonic input-output mappings. }
	\label{fig:basic_lattice}

\end{figure} 
\fi

Integrating a DLN network into ARGate+ leads  to a new architecture called ARGate-L in Fig.~\ref{fig:lattice}. In general, DLNs consist of three types of layers: calibrators, linear embeddings, and lattices, all of which can be constrained to obey partial or full monotonicity between selected inputs and outputs. Linear embedding layers map the inputs linearly to the outputs. Monotonicity between a subsest of inputs/outputs can be forced by choosing non-negative coefficients between them. A lattice is a linearly interpolated multi-dimensional look-up table; each output of the lattice can be constrained to be monotonic in a subset of the inputs. Calibrators are 1-d lattices and can nonlinearly transform a single input and may be used for pre-processing and normalization between layers in the DLN.

The proposed lattice network is illustrated in Fig.~\ref{fig:lattice} for the general case of mapping $K$ auxiliary path losses to the corresponding $K$ fusion targets. In its most general form, this can be done by having $K$ independent subnetworks, one for each fusion target. While a particular architecture is chosen for each subnetwork in Fig.~\ref{fig:lattice}, in practice, it can be simplified/optimized to suit a given application. The key idea in the proposed lattice network design is to ensure the partial monotonicity between each pair of auxiliary loss and fusion target; there are  $K$ such constraints. In the lattice network of Fig.~\ref{fig:lattice}, monotonic inputs/outputs are processed by a calibrator in dashed green while non-monotonic ones are processed by a calibrator in solid gray.  The output of a component is monotonic if any of its inputs is monotonic to ensure the end-to-end monotonicity between a pair of auxiliary loss and fusion targets. Specifically, there exist a green dashed path from each auxiliary loss $k$ to its corresponding fusion target $k$. Imposing these monotonicity constraints acts as a regularization mechanism, making the robust end-to-end learning of fusion weight targets possible and leading to improved performance as we demonstrate later. 

\section{Experimental Settings}

We perform comprehensive comparison of a non-gating fusion CNN baseline \cite{chen2017multi}, the NetGated architecture \cite{patel2017sensor}, and variants of the proposed ARGate architecture on the HAR  \cite{anguita2013public}, driver  identification  \cite{kwak2016know}, and KITTI \cite{Geiger2012CVPR} datasets.

%The mini-batch scheme size is 16 for the HAR dataset, and 128 for the driver identification dataset. We train these networks with 200 epochs for the HAR dataset and 100 epochs for the driver identification dataset.
%For the HAR and the driver identification dataset, an ADAM optimizer is utilized with a learning rate of 0.001. 
%The cross-entropy loss is chosen as the loss for the CNN baseline, the NetGated architecture, and the main model and auxiliary paths of ARGate. 
%For KITTI dataset, an ADAM optimizer is adopted with an initial learning rate of 0.0001, which decays every 30K iterations with 0.8 as a decay factor. All simulations are done with Python 3.6, Pytorch 0.4.0\cite{paszke2017automatic} for HAR and driver identification datasets, and Tensorflow 1.3.0\cite{tensorflow2015-whitepaper} for KITTI dataset using a NVIDIA Tesla K80 GPU.

\subsection{Datasets}

\subsubsection{The Human Activity Recognition Dataset}

The Human Activity Recognition (HAR) dataset~\cite{anguita2013public} 
includes six activities to be recognized: walking, walking upstairs, walking downstairs, sitting, standing, and laying. From accelerometer and gyroscope, nine sensory inputs are utilized for experiments, where each sensory input is distributed between $[-1, 1]$. There are 7,352 examples for training and 2,947 examples for testing.

\subsubsection{The Driver Identification Dataset}

The driver identification dataset \cite{kwak2016know} consists of 10 drivers' cruising data collected by a CarbigsP OBD-II scanner. There are 51 features in total, but 15 features are used in our experiments which is same experimental setup from the original paper. % 51 features were extracted every 1 second while driving out of which  15 features are used in our experiments.
%are: \textit{Long term fuel trim bank1, intake air pressure, accelerator pedal value, fuel consumption, friction torque, maximum indicated engine torque, engine torque, calculated load value, activation of air compressor, engine coolant temperature, transmission oil temperature, front wheel velocity-left and right hand, rear wheel velocity-left hand}, and \textit{torque converter speed}. 
75,501 training and 18,879 testing examples are used. 
%To demonstrate sensor fusion, we treat each of the 15 input features as a sensory input. 
%We have two sensor failure setups. The first one introduces failures onto the five extracted input features.  The second one more physically models failures in terms of raw RGB and depth input data. The corrupted raw RGB/depth data are passed onto feature extraction before feeding a neural network as sensory inputs.  
%The ``new person" setting of \cite{sung2012unstructured} is adopted to  use the data from three people for training and that from the remaining person for testing.

\subsubsection{KITTI Dataset}

In the KITTI dataset \cite{Geiger2012CVPR}, there are two sensory inputs: a RGB image and a velodyne laser scanned bird eye view (BEV) image. The provided 7481 training frames are split into a training and a validation set. For evaluation with detecting the \textit{car} class in images, we adopt the easy, moderate and hard difficulty-level settings provided by KITTI. % Detailed information for the difficulties are demonstrated in Table. ~\ref{table:Three difficulties of performance measurements in KITTI dataset}.

\iffalse

\begin{table}[htbp]

\begin{center}
%\begin{tabular}{|p{1.4cm}|c{1.1cm}|l|l|l|}
\begin{tabular}{|l|l|l|l|}
\hline
\multirow{2}{*}{Difficulty} & Min bounding   & Max & Max \\ 
 & box height  & occlusion & truncation\\ \hline
Easy  & 40 pixels & Fully visible & 15\% \\ \hline
Moderate  & 25 pixels & Partly occluded & 30\% \\ \hline
Hard  & 25 pixels & Difficult to see & 50\% \\ \hline

\end{tabular}
\end{center}
\caption{Three difficulties of performance measurements in KITTI dataset.}
\label{table:Three difficulties of performance measurements in KITTI dataset}
\end{table}

\fi

\subsection{Neural Network Configurations}
\subsubsection{Configurations for the HAR and Driver Identification Datasets}
In the HAR and driver identification datasets, the non-gating CNN baseline, NetGated, and the proposed ARGate variants are compared. The late fusion scheme is utilized in the CNN baseline. For fair comparison, tunable parameters of all neural networks are closely matched. % Details are included in Supplementary Material.
%Fellows are the network setting we are using. \\
%All models are trained using 200 epochs with a batch size of 16 for the HAR dataset, 100 epochs with a batch size of 128 for the driver identification dataset, and 120,000 epochs with a batch size of 1 for KITTI dataset.

\subsubsection{Configurations for the KITTI Dataset}

\textbf{AVOD Baseline.}
We use the Aggregate View Object Detection (AVOD) approach \cite{ku2017joint} as a baseline. 

\begin{figure}[h]

\centering
  \includegraphics[width=0.7\linewidth]{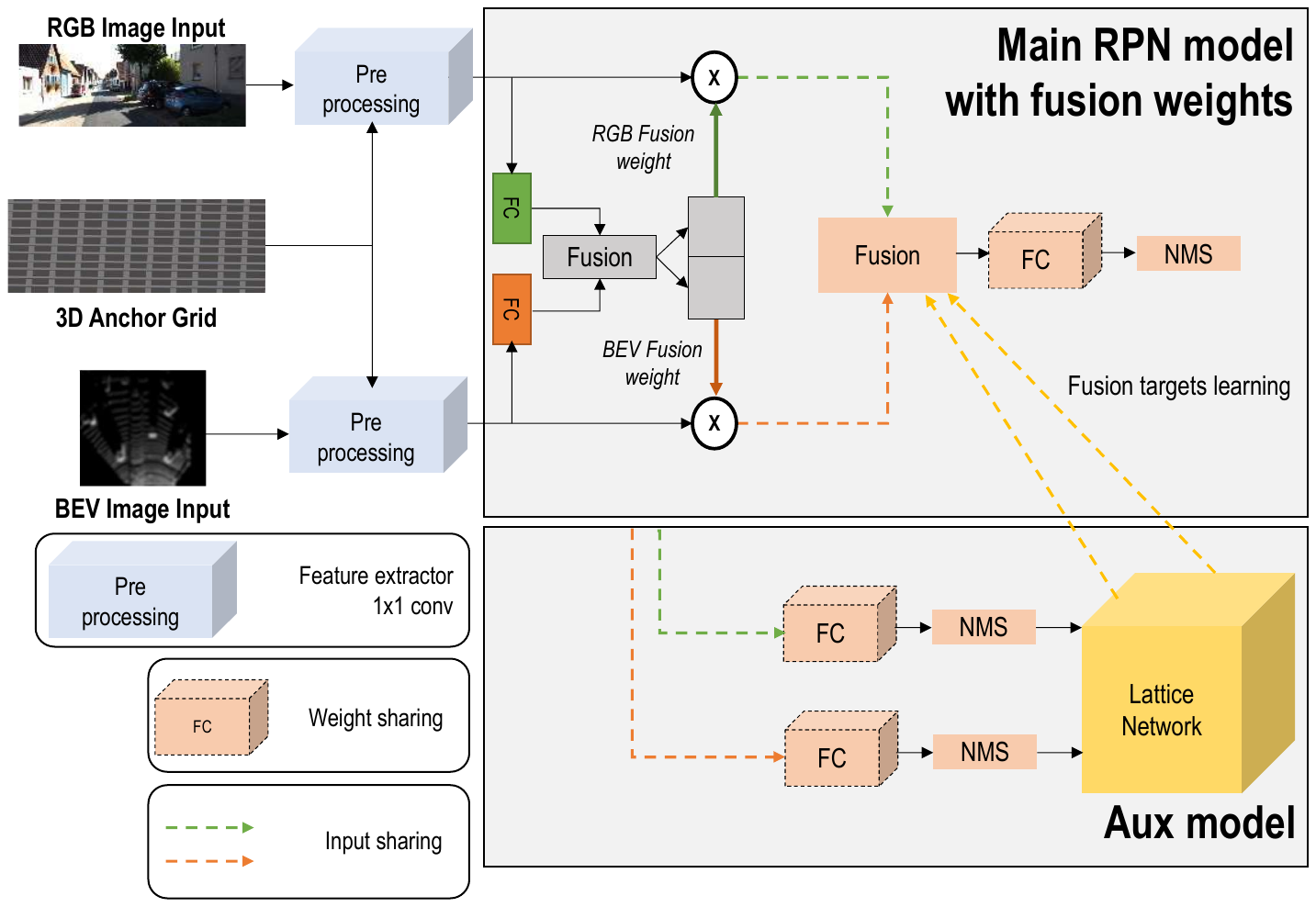}

  \caption{The proposed ARGate architecture with RPN model for training.}
  \label{fig:auxiliary_kitti}

\end{figure}

\textbf{AVOD-ARGate.}
%We embed our ARGate scheme into the AVOD, which has two fusion stages in the region proposal network (RPN) and second stage detection network, respectively. 
% Unlike the input-level fusion adopted in the other two datasets, \textit{ROI-level} fusion is introduced for the KITTI dataset. Each pre-processed ROI from the crop/resize layer is handled by the subsequent set of FC layers.  
We embed our ARGate-L into the AVOD, specifically in the region proposal network(RPN) as shown in Fig.~\ref{fig:auxiliary_kitti}. Feature maps from each input are passed into small ARGate block to produce fusion weights which are regularized by a lattice network.
Outputs from these FC layers are added and passed onto the \textit{fusion} block (a FC layer) to create the fusion weights. These fusion weights are multiplied with the corresponding feature maps from the outputs of the 1x1 feature extracting layers. Furthermore, the fusion weights from RPN is utilized again as gating factors for each modality in the second stage fusion(AVOD) to fortify the network robustness. For target fusion weight learning, the lattice architecture (Cal-Lin-Cal-Lat) is used with the linear embedding layers processing two input channels and one lattice. The total number of tunable parameters of the AVOD baseline and AVOD-ARGate are closely matched for fair performance comparisons. Moreover, since AVOD based algorithms need plane data for training and test, we utilize plane data provided by AVOD for training set, and custom plane data for test set for fair comparison.

% For ARGate-L, the lattice architecture (Cal-Lin-Cal-Lat-Cal-Lin) is used with the linear embedding layers processing two input channels and one lattice.
%And, the shared FC layers are connected to output layers for region proposals.

% The detailed configuration of AVOD-ARGate is included in the Supplementary Materials. The total number of tunable parameters of the AVOD baseline and AVOD-ARGate are closely matched for fair performance comparisons. 

% \subsection{Fusion Weight Normalization} \label{fusion_weights_norm}
% In the original NetGated architecture \cite{patel2017sensor}, the scalar fusion weights are extracted from the FC layer ``FC-con" (Fig.~\ref{fig:NetGated}) without normalization. As a result, the range of the extracted fusion weights is not fully controlled. In this work, we apply fusion weight normalization for both NetGated and the proposed architectures since it provides performance improvements as follows. The fusion weights $\bm w_{fusion}$ outputted by the ``FC-con" layer are normalized by L2 and then $softmax$ normalization: $\bm w_{fusion, n} = softmax \left(l^2_{norm}(\bm w_{fusion})\right)$, where the $l^2_{norm}$ normalizes fusion weights $\bm w_{fusion}$ to be within $[-1,1]$, which are then normalized by $softmax$ to fall in  $[0,1]$ and sum up to $1.0$. As such, each scalar fusion weight can be readily interpreted as a measure of criticality of the corresponding modality for the fusion task. 

\subsection{Sensor Failures} \label{Sensor Noise and Failures}
Apart from using the clean data in the HAR and driver identification datasets, we introduce sensor failures and set up various training/testing sets to comprehensively compare the robustness/generalization of different architectures.

\subsubsection{Modeling of failing sensors} 

All clean scalar inputs are normalized to be within $[-1, 1]$. 
We use two schemes: \emph{uniform}, and \emph{Gaussian}, to model a failing sensor by setting its input values respectively to pure noise following a uniform distribution between $[-1, 1]$, and pure noise following the Gaussian distribution $\mathcal{N}(0, 1)$.  

\subsubsection{Corrupted examples for training/testing}

\label{Dynamic_sensor_failure_channels}

For HAR and driver identification datasets, we use clean and corrupted examples in both the training and test sets, where in each set $\frac{1}{3}$ of the examples are randomly chosen and kept clean while the remaining ones are corrupted by one or more failing sensors using one of the approaches described below.   
 
\textbf{Fixed Failing Sensor Assignment.} This mimics the situation in which a number of sensors have failed permanently. We select $n_{fclean}$ channels out of a total of $n$ sensors to be clean and assume that all remaining sensors have permanent failure when setting up the corrupted examples of for both the training and test sets in each experiment.

\textbf{Random Failing Sensor Assignment.} To  more closely mimic random nature of sensor failures, for each corrupted example in the training/test set, we randomly select $n_{rclean}$ channels out of all $n$ sensors to be clean and corrupt the remaining channels. As such,  sensors that have failed may vary from one example to another.  $n_{fclean}$ and $n_{rclean}$  are varied to for different severity levels of failures. 

\textbf{Failing Sensor Generation Test.} This tests model generation by using a test set containing corrupted examples which have a larger or different number of failing sensors from the corrupted examples used to train the model, i.e. the test set has examples with a severe level of sensor failure.  

% \subsection{Batch Loss}

% A cross-entropy loss function is applied to calculate classification loss in the three architectures, each element in (\ref{three_term_loss}) is calculated example by example.

% \begin{center}
% \begin{equation}\label{norm}
% \begin{split}
%     Loss_{batch} = \frac{1}{n}\sum_{i=1}^{n}({{Loss}_{main_i}} + \\ \alpha({{W}_{fusion_i}-\reallywidehat{({e}^{-{Loss}_{aux_i}^2}}))^2} \\ 
%     + \beta{{W}_{fusion_i}{{Loss}_{aux_i}}})
% \end{split}
% \end{equation}
% \end{center}

% In (\ref{norm}), note that $n$ represents batch-size, and  $i$ is the $i_{th}$ example of a current mini-batch. $Loss_{aux}$ is exploited for giving feedback to fusion stage in the FC layer "FC-con", as a red line shown in Fig. \ref{fig:aux_loss}. 

%------------------------------------------------------------------------

\section{Evaluation}\label{sec:results}

\label{Experimental_results}

%In driver identification dataset, the feature extractor parts are composed by 3x3x8 same padding convolution layer, 2x2 max pooling layer, 3x3x4 same padding convolution layer and finally a 2x2 max pooling layer. For the classification parts we apply two fully connected layers, to get 600 dimension output first then down to 200. Finally, we use another fully connected layer to generate 12 output then apply Softmax layers to them. Each auxiliary path shares the feature extractor parts of the main model(not only the setting of network but also the weights) and for classification parts, the networks setting are same as the main model. 

%-------------------------------------------------------------------------------------

%----------------------------------------------------

\begin{figure*} 
	\centering
	\includegraphics[width=1.0\linewidth]{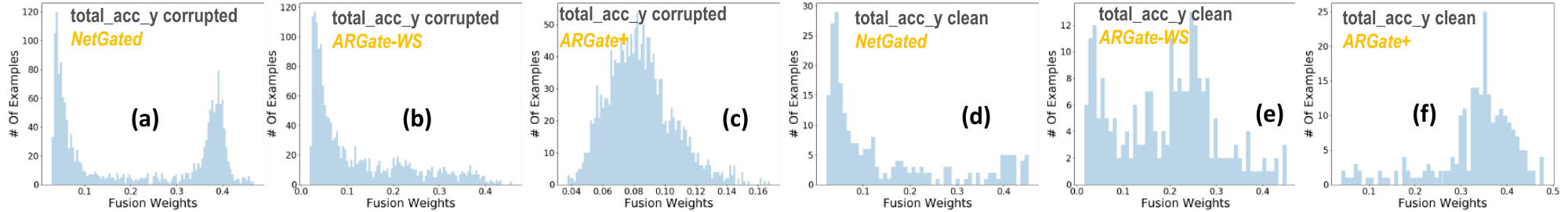}

	\caption{Fusion weight distributions of the clean or corrupted channel total\_acc\_y extracted by NetGated, ARGate-WS and ARGate+ under random failing sensor assignment with $n_{rclean} =1$. (a),(b) and (c) show the fusion weights distributions of the NetGated, ARGate-WS  and ARGate+ models, respectively, when total\_acc\_y is corrupted. (d),(e) and (f) are the distributions of the NetGated, ARGate-WS and ARGate+ models, respectively, when total\_acc\_y is clean.}
	\label{fig:Weight_distribution}
	
\end{figure*}

\begin{table*}[t]

\begin{center}
%\begin{tabular}{|p{1.4cm}|c{1.1cm}|l|l|l|}
\begin{tabular}{|c|c|l|l|l|l|l|l|}
\hline
\# Clean & Failure  & \multirow{2}{*}{Baseline}  & \multirow{2}{*}{NetGated} & ARGate-& AR & AR\\%\multirow{2}{*}{ARGate+} \\ 
Channels & Model &  & & WS & Gate+ & Gate-L\\ \hline\hline%\multirow{2}{*}{AR} \\ Channels & Model &  & &-WS & WS-FWR & Gate+\\ \hline
All Clean & - & 94.06 & 94.50 & 94.96  &95.69 & \textbf{96.71}\\ \hline
      
$n_{rclean}$=8 & Uniform & 92.35 & 92.20 & 92.45  & 92.57 & \textbf{94.06}\\ \cline{2-7}
      & Gaussian & 92.94 & 93.28 & 94.97  & 94.13 & \textbf{94.81}\\ \hline
      
$n_{rclean}$=5 & Uniform & 86.73  & 86.80 & 88.53 & 87.68 & \textbf{88.77}\\ \cline{2-7}
      & Gaussian & 88.41 & 89.04 & 89.52  & 90.12 & \textbf{91.35}\\ \hline
      
$n_{rclean}$=1 & Uniform &  62.06 & 62.90 & 65.69  & 67.09 & \textbf{69.63}\\ \cline{2-7}
      & Gaussian & 69.67 & 70.54 & 71.83 & \textbf{73.19} & 72.78\\ \hline

\end{tabular}
\end{center}

\caption{Prediction accuracies(in \%) under clean data and random failing sensor assignment for the HAR dataset.}
\label{HAR_table}

\end{table*}

%----------------------------------------------------
%-------------------------------------------------------------------------------------------
%In Table.~\ref{HAR_table_compare_ARGate-WS_ARGate-F} and Table.~\ref{CAD_60_table_compare_ARGate-WS_ARGate-F}, ARGate is verified in different sensor failure cases with dynamic noisy channels. ARGate-WS-FWR shows an improvement in accuracy over ARGate-WS, which illustrates that regularizing fusion weights with $Loss_{aux}$ provides a positive impact to the performance. The improvement achieved through ARGate-F shows that multiplying fusion weights to $Loss_{aux}$ can help avoid problems of \ref{three_term_loss_without_Fusion_weights}. We use (\ref{three_term_loss_without_Fusion_weights}) and (\ref{three_term_loss}) for the loss functions in ARGate-WS and ARGate-F in our following experiments.
%ARGate is verified in different sensor failure cases with dynamic noisy channels. ARGate-WS-FWR improves the performance comparing to ARGate-WS, which represents that regularizing fusion weights through $Loss_{aux}$ helps the neural network to avoid the problem in loss function (\ref{three_term_loss_without_Fusion_weights}). We compare our proposed architectures ARGate-WS, ARGate-WS-FWR, and ARGate-F in experiments.

\subsection{Quality of Fusion Weight Extraction}

To shed some light on the fusion mechanisms of the NetGated, ARGate-WS and the proposed ARGate+ architectures, we examine the distributions of the trained fusion weight of the sensory input  $total\_acc\_y$ under the HAR dataset. If the  $total\_acc\_y$ is corrupted, then the fusion weight of itself should be distributed lower than the fusion weight of clean $total\_acc\_y$. Please note that ARGate-WS only utilizes weight sharing so that we compare our proposed ARGate+ with ARGate-WS in terms of fusion weight regularization. The sensor failures are modeled using the random failing sensor assignment with uniform distribution and with  $n_{rclean} =1$ so that 8 out of 9 sensory inputs are corrupted in each example. To show the effects of sensor failures on the fusion weights, we split the examples into two subsets: one in which $total\_acc\_y$ is corrupted with 8 other inputs and the other subset where  $total\_acc\_y$ in the only clean input. Fig. 7(a, b, c) displays the fusion weight distributions of the first subset for the three architectures while Fig. 7(d, e, f)  shows those of the second subset. 

%In the ARGate-WS loss function model in (\ref{two_term_loss}), $\alpha$ is set as 5. For ARGate+ model in (\ref{three_term_loss}) $\alpha$ and $\beta$ are set to 9 and 5, respectively. Same two-step normalization in Section \ref{fusion_weights_norm} is implemented to normalize ${e}^{-{Loss}_{aux_i}^2}$ in (\ref{three_term_loss}). Since $W_{fusion}$ is normalized by softmax, summation of fusion weights of 9 features should be 1.

One may expect that, in a properly trained model, the fusion weight value for a corrupted sensory modality shall be much smaller than when the modality is clean. However even when $total\_acc\_y$ is corrupted,  Fig. 7(a, b, c) show that the $total\_acc\_y$ fusion weight distribution of the NetGated architecture has a peak around the large value of 0.4 which is not presented in the case of ARGate-WS. The distribution of ARGate-WS has a much reduced mass on large fusion weight values compared to that of the NetGated. Furthermore, the fusion weight value can go beyond 0.4 in NetGated and ARGate-WS while it is pretty much constrained between 0.06 and 0.1 in ARGate+. 
One also expects that the fusion weight of $total\_acc\_y$ shall be large for the second subset since  $total\_acc\_y$ is the only clean input.  However, as seen in Fig. 7d, the distribution of NetGated has a large population mass on low fusion weight values. The population mass on low fusion weight values gets reduced significantly in the case of ARGate-WS Fig. 7e, which is further reduced in ARGate+ for which  most fusion weights are distributed between 0.33 and 0.45 as shown in Fig. 7f.

% Under this setup, the prediction accuracies of the NetGated, ARGate-WS, ARGate+, and ARGate-L models are 62.90\%, 65.69\%, 67.09\%, and 67.59\%, respectively in Table. \ref{HAR_table}. We expect that the performance improvements of the ARGate architectures may be attributed to their improved quality in fusion weight target learning. 

% Based on the weight distributions presented in Fig.~\ref{fig:Weight_distribution}, we expect that weight sharing (WS) employed  in ARGate-WS makes it more robustly learn the quality of each sensory input to achieve a higher accuracy than NetGated. And the inclusion of fusion weight regulation (FWS) and auxiliary loss weighting (ALW)  in ARGate+ and lattice in ARGate-L further improve the learning quality of fusion weights, thereby producing the best classification result among the three architectures. 

\begin{table}[htbp]
\begin{center}
%\begin{tabular}{|p{1.4cm}|c{1.1cm}|l|l|l|}

\begin{tabular}{|c|l|l|l|l|}
\hline
\# Clean & \multirow{2}{*}{Baseline}  & \multirow{2}{*}{NetGated} & AR & AR \\ 
Channels  &  & & Gate+ & Gate-L \\ \hline\hline
$n_{fclean}$=6  & 87.68 & 89.28 & 91.01 & \textbf{92.97}\\ \hline
$n_{fclean}$=5  & 80.59 & 81.94 & 84.52 & \textbf{89.01}\\ \hline

\end{tabular}
\end{center}

\caption{Prediction accuracies(in \%) under fixed failing sensor assignment for the HAR dataset.}
\label{Accuracy on sensor failure with fixed noise channel in HAR dataset}

\end{table}

\subsection{Results on the HAR Dataset} \label{HAR_Results}

\textbf{[Fixed Failing Sensor Assignment]}
%\label{Simulation Results of Sensor Failure with Fixed Noisy Channel}
%\label{Sensor_Failure_scheme}
We consider two cases where the number of clean input channels ${n}_{fclean}$ is set to $5$ and $6$, respectively. When ${n}_{fclean}=5$, $body\_total\_acc\_x$ $body\_acc\_x$ and  $body\_gyro\_x$ are corrupted and set to uniform-ally distributed noise while $body\_acc\_z$ and $body\_gyro\_x$ are corrupted by  uniform-ally distributed noise when ${n}_{fclean}=6$.  Table~\ref{Accuracy on sensor failure with fixed noise channel in HAR dataset} shows that our proposed ARGate architectures outperform the CNN  and NetGated significantly while adopting the lattice network in ARGate-L further improves over ARGate+.

\textbf{[Random Failing Sensor Assignment]} In Table~\ref{HAR_table} compares the baseline CNN, NetGated, ARGate-WS, proposed ARGate+, and ARGate-L architectures with the number of randomly chosen clean sensors $n_{rclean}\in\{1,5,8\}$. 
When all channels are clean, NetGated has 0.44\% prediction accuracy improvement over the baseline while ARGate+, and ARGate-L outperform the baseline by 1.63\%, and 2.65\%, respectively. ARGate architectures always have better accuracy than the baseline and NetGated, and in general ARGate+ further improves over ARGate-WS which only employs weight sharing (WS) between the main and auxiliary model, demonstrating the effect of fusion weight regularization with auxiliary losses (FWR). ARGate-L is the best-performing model, which incorporates WS, FWR, and the lattice network for monotonic fusion weight target learning. %(Fig.~\ref{fig:complete_argate}). 
With $n_{rclean}=1$ and sensor failures which are modeled using uniformly distributed noise, ARGate-L outperforms the baseline, NetGated, ARGate-WS, and ARGate by 7.57\%,  6.73\%,  3.94\%, and 2.54\%, respectively. We expect that the performance improvements of the ARGate architectures may be attributed to their improved quality in the fusion weight target learning.

\begin{table}[htbp]

\begin{center}
%\begin{tabular}{|p{1.4cm}|c{1.1cm}|l|l|l|}

\begin{tabular}{|c|l|l|l|l|}
\hline
\# Clean & \multirow{2}{*}{Baseline}  & \multirow{2}{*}{NetGated} & AR & AR \\ 
Channels  &  & & Gate+ & Gate-L \\ \hline\hline
(1,2)(3,8)  & 72.91 & 72.75  & 77.10  & \textbf{77.16}\\ \hline
(1,3)(4,8)  & 70.98 & 70.78  & 75.43  & \textbf{76.15}\\ \hline
(1,4)(5,8)  & 69.38 & 69.53  & 73.06  & \textbf{73.64}\\ \hline

\end{tabular}
\end{center}

\caption{Prediction accuracies(in \%) under failing sensor generation test for the HAR dataset.}
\label{HAR under general noise schemes all combine}
\end{table}

\begin{table}[htbp]

\begin{center}
%\begin{tabular}{|p{1.4cm}|c{1.1cm}|l|l|l|}
\resizebox{.8\columnwidth}!{ }
\begin{tabular}{|c|l|l|l|l|}
\hline
\# Clean & \multirow{2}{*}{Baseline}  & \multirow{2}{*}{NetGated} & AR & AR       \\ 
Channels & & & Gate+ & Gate-L \\ \hline\hline
$n_{fclean}$=5  & 79.48  & 81.46  & 82.29  & \textbf{87.73}  \\ \hline
$n_{fclean}$=7  & 90.39  & 92.81  & 94.56  & \textbf{95.34}  \\ \hline
\end{tabular}
\end{center}

\caption{Prediction accuracies(in \%) under fixed failing sensor assignment for the driver identification dataset.}
\label{Accuracy on sensor failure with fixed noise channel in driver identification dataset}
\end{table}

\begin{table}[htbp]

\begin{center}
\scalebox{0.83}{
\begin{tabular}{|c|c|l|l|l|l|l|l|}
\hline
\# Clean & Failure  & \multirow{2}{*}{Baseline}  & \multirow{2}{*}{NetGated} & AR & AR \\%\multirow{2}{*}{ARGate+} 
Channels & Model &  &  & Gate+ & Gate-L \\ \hline\hline
All Clean & - & 95.49 & 95.79 & 96.65 & \textbf{96.78} \\ \hline
      
$n_{rclean}$=12 & Uniform &  {93.43} &  {92.19} & {93.50} & \textbf{93.58}\\ \cline{2-6}
      & Gaussian & {92.19} & {91.40} &{92.31} & \textbf{93.16}\\ \hline
      
$n_{rclean}$=8 & Uniform & 83.50 & 80.61  & 85.08 & \textbf{86.86}\\ \cline{2-6}
      & Gaussian & 80.82 & 79.19 &82.23 & \textbf{85.48} \\ \hline
      
$n_{rclean}$=5 & Uniform & 65.36 & 62.21 & 68.96 & \textbf{73.85} \\ \cline{2-6}
      & Gaussian & 65.84 & {62.69} & {68.20} &\textbf{76.08} \\ \hline
      
%One Clean & \multirow{2}{*}{Large} & \multirow{2}{*}{58.08} & \multirow{2}{*}{57.99} & \multirow{2}{*}{61.08} \\ \hline

\end{tabular}}
\end{center}

\caption{Accuracies(in \%) under clean data and random failing sensor assignment for the driver identification dataset.}
\label{driver identification_table}

\end{table}

\textbf{[Failing Sensor Generation Test]}
In Table~\ref{HAR under general noise schemes all combine},  the first column specifies the numbers of randomly chosen failing channels used in the training and test sets. For example, (1,2)(3,8) means that the number of failing sensors for training are randomly picked from [1,2]  while the range of the failing sensors for testing is [3,8]. Essentially, we evaluate the generalization of the models by including corrupted examples with more failing channels in the test set than in the training set.  Failing sensors are modeled using uniformly distributed noise.  NetGated can underperform the baseline while ARGate-L always outperforms  the baseline and NetGated by upto $5.17\%$ and $5.37\%$, respectively.

\subsection{Ablation studies on HAR Dataset}
Table.~\ref{HAR_table_ablation} shows the effect of components of the ARGate architectures on prediction performance. The results of Netgated is shown in this table as gate only since weight sharing (WS), fusion weight regularization (FWR), and fusion target learning(+ and L) are not applied. ARGate-WS is composed of auxiliary paths, WS and normalization of fusion weights in the main model. As we can see, WS and the normalization of fusion weights helps overall performance, especially when it comes with severe noisy conditions. When FWR module is added to ARGate-WS, the regularization technique makes the ARGate architecture robust and improves performance. Lastly, From Table.~\ref{HAR_table}, thanks to the monotonic fusion target learning scheme (ARGate-L), the improvement of prediction performance is bigger than when the ARGate is composed with WS or FWR. 

\begin{table}[t]

\begin{center}
%\begin{tabular}{|p{1.4cm}|c{1.1cm}|l|l|l|}
\begin{tabular}{|c|c|l|l|l|l|l|l|}
\hline
\# Clean & Failure  & \multirow{2}{*}{Gate only} & ARGate-& ARGate- \\%\multirow{2}{*}{ARGate+} \\ 
Channels & Model &  & WS &WS-FWR \\ \hline\hline%\multirow{2}{*}{AR} \\ Channels & Model &  & &-WS & WS-FWR & Gate+\\ \hline
All Clean & -  & 94.50 & 94.96 & 95.09  \\ \hline
      
$n_{rclean}$=8 & Uniform  & 92.20 & 92.45 & 92.46  \\ \cline{2-5}
      & Gaussian  & 93.28 & 94.97 & 94.35  \\ \hline
      
$n_{rclean}$=5 & Uniform   & 86.80 & 88.53 & 89.17 \\ \cline{2-5}
      & Gaussian & 89.04 & 89.52 & 90.07  \\ \hline
      
$n_{rclean}$=1 & Uniform  & {62.90} & {65.69} & {66.09} \\ \cline{2-5}
      & Gaussian & {70.54} & {71.83} & {72.58} \\ \hline

\end{tabular}
\end{center}

\caption{A comparison of performance with different components under clean data and random failing sensor assignment for the HAR dataset.}
\label{HAR_table_ablation}

\end{table}

\subsection{Results on the Driver Identification Dataset} 
\label{driver-identification_Results}

Based on the results above, our proposed ARGate+ and ARGate-L architectures are examined with the driver identification dataset.

\textbf{[Fixed Failing Sensor Assignment]}
In Table~\ref{Accuracy on sensor failure with fixed noise channel in driver identification dataset},  the corrupted inputs are set to uniformly distributed noise. When $n_{fclean}=5$, \textit{Long Term Fuel Trim Bank1, Maximum indicated engine torque, Calculated LOAD value, Activation of Air compressor}, and \textit{Engine coolant temperature} are corrupted. When $n_{fclean}=7$, two more input channels, \textit{Intake air pressure} and \textit{Fuel consumption} are corrupted. ARGate-L significantly outperforms the other two models. When $n_{fclean}=5$, ARGate-L improves over the baseline and NetGated by 8.25\% and 6.27\%, respectively.

\textbf{[Random Failing Sensor Assignment]} In Table~\ref{driver identification_table}, different models with the number of randomly selected clean channels $n_{rclean}\in\{5,8,12\}$ are compared.  When all 15 input channels are clean, the proposed ARGate-L improves the baseline and NetGated by 1.29\% and 0.99\%, respectively. In many cases, NetGated is worse than the baseline. ARGate-L always has the best performance among all models. For example, with $n_{rclean}=5$ and the uniform noise sensor failure model, ARGate-L significantly outperforms the baseline and NetGated by 8.49\% and 13.39\%, respectively.

\textbf{[Failing Sensor Generation Test]}
We  compare the generalization of two models in Table~\ref{driver identification under general noise schemes all combine} with the same notation of 
Table~\ref{HAR under general noise schemes all combine}.  
%(2,4)(1,4) means in training phase, the number of noisy channel are randomly selected among integers between [2,4], ($n_{rclean}\in\{2,3,4\}$). The number of noisy channel during testing are randomly selected among integers between [1,4], ($n_{rclean}\in\{1,2,3,4\}$). %To simulate testing cases are more complicated than training. Simulations are done with failed inputs under uniform distribution.
ARGate-L demonstrates noticeable improvements over all other  models.

\begin{table}[htbp]

\begin{center}
\begin{tabular}{|l|l|l|l|l|}
\hline
%\multicolumn{1}{|c|}{# Failing Channels} & \multicolumn{1}{c|}{Baseline}       & %\multicolumn{1}{c|}{NetGated}   & \multicolumn{1}{c|}{ARGate+}        \\ \hline
\# Failing  & \multirow{2}{*}{Baseline}  & \multirow{2}{*}{NetGated} & AR & AR       \\ 
Channels & & & Gate+ & Gate-L \\ \hline\hline
(1,2)(3,15)                          & \multicolumn{1}{c|}{60.38} & 
\multicolumn{1}{c|}{60.39} &
\multicolumn{1}{c|}{59.42} & \multicolumn{1}{c|}{\textbf{64.16}} \\ \hline

\end{tabular}
\end{center}

\caption{Prediction accuracies(in \%) under the failing sensor generation test for the driver identification dataset.}
\label{driver identification under general noise schemes all combine}
\end{table}

\subsection{Results on the KITTI Dataset}

\label{KITTI_Results}
We compare our ARGate-L architecture to the AVOD baseline \cite{ku2017joint} on car detection in KITTI validation and test set. Average Precision (AP) in 2D image frame, oriented overlap on image, AP in BEV, and AP in 3D metrics are used for performance comparison. 

In terms of validation results in Table. ~\ref{table:AP_KITTI_val}, 1.17\% improvement is made on 3D car detection benchmark in moderate difficulty. Furthermore, about 0.4\% of improvements are found from BEV AP. With fusion weights utilized on both RPN and AVOD network, our ARGate-L architecture improves the detection performance in the validation set. 

For test results evaluated by official KITTI online server, since the AVOD plane data is not provided for test set, we use our custom plane data for AVOD and ARGate-L for fair comparison. Based on this setup, in the moderate difficulty, our ARGate architecture improves 4.63\% in 2D car detection benchmark. For orientation, 4.32\% improvement is observed. Lastly, for 3D detection and BEV, the proposed architecture outperforms AVOD by 4.81\% and 2.22\%, respectively. Furthermore, similar range of improvements are shown on the hard difficulty, which shows strength of the proposed architecture, especially on touch situation. Overall, our proposed techniques outperform the baseline AVOD rather noticeably. 

\begin{table}[htbp]

\begin{center}
\scalebox{0.83}{
\begin{tabular}{|c|c|c|c|c|}
\hline
\multirow{2}{*}{Network} & \multirow{2}{*}{Benchmark}  & \multirow{2}{*}{Easy}  & \multirow{2}{*}{Moderate} & \multirow{2}{*}{Hard}  \\%\multirow{2}{*}{ARGate+} 
 &  &  &  &   \\ \hline\hline

\multirow{2}{*}{AVOD} 
      & Car (3D Detection) & {84.41} & {74.44} &{68.65} \\ \cline{2-5}
      & Car (BEV) & {89.72} & {86.85} &{79.69} \\ \hline
      
\multirow{2}{*}{AVOD-ARGate} 
      & Car (3D Detection) & \textbf{84.61} & \textbf{75.61} & {68.65} \\ \cline{2-5}
      & Car (BEV) & \textbf{89.95} & \textbf{87.23} &\textbf{79.89} \\ \hline

%One Clean & \multirow{2}{*}{Large} & \multirow{2}{*}{58.08} & \multirow{2}{*}{57.99} & \multirow{2}{*}{61.08} \\ \hline

\end{tabular}}
\end{center}

\caption{Average Precision (in \%) comparison of car detection on the KITTI \textit{validation} set.}
\label{table:AP_KITTI_val}

\end{table}

\iffalse
\begin{table}[htbp]
\begin{center}
    \begin{tabular}{|l|l|l|l|}
    \hline
   Network & Easy & Moderate & Hard \\ \hline\hline
    AVOD & 84.41 & 74.44 & 68.65    \\ \hline
    AVOD-ARGate & \textbf{84.61} & \textbf{75.61} &  \textbf{68.65}    \\ \hline
    \end{tabular}
\end{center}
\caption{Average Precision (in \%) comparison of car detection on the KITTI \textit{validation} set.}
\label{table:AP_KITTI_val}
\end{table}
\fi

% based on the KITTI's validation set.
% Our ARGate architecture outperforms the baseline AVOD by 0.2\% for easy, and 0.57\% for the moderate difficulty in the \textit{validation} set under the matched number of parameters in neural network architectures.
 
\begin{table}[htbp]

\begin{center}
\scalebox{0.83}{
\begin{tabular}{|c|c|c|c|c|}
\hline
\multirow{2}{*}{Network} & \multirow{2}{*}{Benchmark}  & \multirow{2}{*}{Easy}  & \multirow{2}{*}{Moderate} & \multirow{2}{*}{Hard}  \\%\multirow{2}{*}{ARGate+} 
 &  &  &  &   \\ \hline\hline

\multirow{4}{*}{AVOD} & Car (Detection) &  {90.17} &  {79.77} & {74.84} \\ \cline{2-5}
      & Car (Orientation) & {89.96} & {79.19} &{74.16} \\ \cline{2-5}
      & Car (3D Detection) & {73.32} & {59.74} &{55.08} \\ \cline{2-5}
      & Car (BEV) & {88.06} & {77.80} &{71.16} \\ \hline
      
\multirow{4}{*}{AVOD-ARGate} & Car (Detection) &  \textbf{92.36} &  \textbf{84.40} & \textbf{79.59} \\ \cline{2-5}
      & Car (Orientation) & \textbf{92.06} & \textbf{83.51} & \textbf{78.63} \\ \cline{2-5}
      & Car (3D Detection) & \textbf{76.52} & \textbf{64.55} & \textbf{60.01} \\ \cline{2-5}
      & Car (BEV) & {87.85} & \textbf{80.02} & \textbf{75.39} \\ \hline

%One Clean & \multirow{2}{*}{Large} & \multirow{2}{*}{58.08} & \multirow{2}{*}{57.99} & \multirow{2}{*}{61.08} \\ \hline

\end{tabular}}
\end{center}

\caption{Average Precision (in \%) comparison of car detection on the KITTI \textit{test} set.}
\label{table:AP_KITTI_test}

\end{table}

%------------------------------------------------------------------------

\section{Conclusion}

We have proposed the ARGate architectures for resilient sensor fusion by addressing the limitations of the conventional fusion schemes including the existing gating architectures. Leveraging the two proposed regularization techniques, namely, fusion weight regularization with auxiliary losses weighting, and monotonic fusion target learning,  the proposed gating architectures incorporate an auxiliary model to regularize the main
model to robustly learn the fusion weight for each modality. 
Our architectures have demonstrated significant performance improvements over other models particularly in the presence of sensor failures.

\bibliographystyle{cas-model2-names}

\bibliography{cas-refs}

\begin{thebibliography}{21}
\expandafter\ifx\csname natexlab\endcsname\relax\def\natexlab#1{#1}\fi
\providecommand{\url}[1]{\texttt{#1}}
\providecommand{\href}[2]{#2}
\providecommand{\path}[1]{#1}
\providecommand{\DOIprefix}{doi:}
\providecommand{\ArXivprefix}{arXiv:}
\providecommand{\URLprefix}{URL: }
\providecommand{\Pubmedprefix}{pmid:}
\providecommand{\doi}[1]{\href{http://dx.doi.org/#1}{\path{#1}}}
\providecommand{\Pubmed}[1]{\href{pmid:#1}{\path{#1}}}
\providecommand{\bibinfo}[2]{#2}
\ifx\xfnm\relax \def\xfnm[#1]{\unskip,\space#1}\fi
%Type = Article
\bibitem[{Ali et~al.(2020)Ali, El-Sappagh, Islam, Kwak, Ali, Imran and
  Kwak}]{ali2020smart}
\bibinfo{author}{Ali, F.}, \bibinfo{author}{El-Sappagh, S.},
  \bibinfo{author}{Islam, S.R.}, \bibinfo{author}{Kwak, D.},
  \bibinfo{author}{Ali, A.}, \bibinfo{author}{Imran, M.},
  \bibinfo{author}{Kwak, K.S.}, \bibinfo{year}{2020}.
\newblock \bibinfo{title}{A smart healthcare monitoring system for heart
  disease prediction based on ensemble deep learning and feature fusion}.
\newblock \bibinfo{journal}{Information Fusion} \bibinfo{volume}{63},
  \bibinfo{pages}{208--222}.
%Type = Inproceedings
\bibitem[{Anguita et~al.(2013)}]{anguita2013public}
\bibinfo{author}{Anguita}, et~al., \bibinfo{year}{2013}.
\newblock \bibinfo{title}{A public domain dataset for human activity
  recognition using smartphones.}, in: \bibinfo{booktitle}{ESANN}.
%Type = Article
\bibitem[{Arevalo et~al.(2017)Arevalo, Solorio, Montes-y G{\'o}mez and
  Gonz{\'a}lez}]{arevalo2017gated}
\bibinfo{author}{Arevalo, J.}, \bibinfo{author}{Solorio, T.},
  \bibinfo{author}{Montes-y G{\'o}mez, M.}, \bibinfo{author}{Gonz{\'a}lez,
  F.A.}, \bibinfo{year}{2017}.
\newblock \bibinfo{title}{Gated multimodal units for information fusion}.
\newblock \bibinfo{journal}{arXiv preprint arXiv:1702.01992} .
%Type = Inproceedings
\bibitem[{Chen et~al.(2017)}]{chen2017multi}
\bibinfo{author}{Chen}, et~al., \bibinfo{year}{2017}.
\newblock \bibinfo{title}{Multi-view 3d object detection network for autonomous
  driving}, in: \bibinfo{booktitle}{IEEE CVPR}, p.~\bibinfo{pages}{3}.
%Type = Article
\bibitem[{Dehzangi et~al.(2017)Dehzangi, Taherisadr and
  ChangalVala}]{dehzangi2017imu}
\bibinfo{author}{Dehzangi, O.}, \bibinfo{author}{Taherisadr, M.},
  \bibinfo{author}{ChangalVala, R.}, \bibinfo{year}{2017}.
\newblock \bibinfo{title}{Imu-based gait recognition using convolutional neural
  networks and multi-sensor fusion}.
\newblock \bibinfo{journal}{Sensors} \bibinfo{volume}{17},
  \bibinfo{pages}{2735}.
%Type = Inproceedings
\bibitem[{Gao et~al.(2019)Gao, You, Zhang, Wang and Li}]{gao2019multi}
\bibinfo{author}{Gao, P.}, \bibinfo{author}{You, H.}, \bibinfo{author}{Zhang,
  Z.}, \bibinfo{author}{Wang, X.}, \bibinfo{author}{Li, H.},
  \bibinfo{year}{2019}.
\newblock \bibinfo{title}{Multi-modality latent interaction network for visual
  question answering}, in: \bibinfo{booktitle}{Proceedings of the IEEE
  International Conference on Computer Vision}, pp.
  \bibinfo{pages}{5825--5835}.
%Type = Inproceedings
\bibitem[{Geiger et~al.(2012)}]{Geiger2012CVPR}
\bibinfo{author}{Geiger, A.}, et~al., \bibinfo{year}{2012}.
\newblock \bibinfo{title}{Are we ready for autonomous driving? the kitti vision
  benchmark suite}, in: \bibinfo{booktitle}{IEEE CVPR}.
%Type = Inproceedings
\bibitem[{Kendall et~al.(2018)Kendall, Gal and Cipolla}]{kendall2018multi}
\bibinfo{author}{Kendall, A.}, \bibinfo{author}{Gal, Y.},
  \bibinfo{author}{Cipolla, R.}, \bibinfo{year}{2018}.
\newblock \bibinfo{title}{Multi-task learning using uncertainty to weigh losses
  for scene geometry and semantics}, in: \bibinfo{booktitle}{Proceedings of the
  IEEE conference on computer vision and pattern recognition}, pp.
  \bibinfo{pages}{7482--7491}.
%Type = Article
\bibitem[{Kim et~al.(2018)}]{kim2018robust}
\bibinfo{author}{Kim}, et~al., \bibinfo{year}{2018}.
\newblock \bibinfo{title}{Robust deep multi-modal learning based on gated
  information fusion network}.
\newblock \bibinfo{journal}{arXiv preprint arXiv:1807.06233} .
%Type = Article
\bibitem[{Ku et~al.(2017)}]{ku2017joint}
\bibinfo{author}{Ku}, et~al., \bibinfo{year}{2017}.
\newblock \bibinfo{title}{Joint 3d proposal generation and object detection
  from view aggregation}.
\newblock \bibinfo{journal}{arXiv preprint arXiv:1712.02294} .
%Type = Inproceedings
\bibitem[{Kwak et~al.(2016)}]{kwak2016know}
\bibinfo{author}{Kwak}, et~al., \bibinfo{year}{2016}.
\newblock \bibinfo{title}{Know your master: Driver profiling-based anti-theft
  method}, in: \bibinfo{booktitle}{2016 14th Annual Conference on Privacy,
  Security and Trust (PST)}, \bibinfo{organization}{IEEE}. pp.
  \bibinfo{pages}{211--218}.
%Type = Article
\bibitem[{Mauldin et~al.(2018)Mauldin, Canby, Metsis, Ngu and
  Rivera}]{mauldin2018smartfall}
\bibinfo{author}{Mauldin, T.R.}, \bibinfo{author}{Canby, M.E.},
  \bibinfo{author}{Metsis, V.}, \bibinfo{author}{Ngu, A.H.},
  \bibinfo{author}{Rivera, C.C.}, \bibinfo{year}{2018}.
\newblock \bibinfo{title}{Smartfall: A smartwatch-based fall detection system
  using deep learning}.
\newblock \bibinfo{journal}{Sensors} \bibinfo{volume}{18},
  \bibinfo{pages}{3363}.
%Type = Inproceedings
\bibitem[{Mees et~al.(2016)}]{mees2016choosing}
\bibinfo{author}{Mees}, et~al., \bibinfo{year}{2016}.
\newblock \bibinfo{title}{Choosing smartly: Adaptive multimodal fusion for
  object detection in changing environments}, in: \bibinfo{booktitle}{IEEE
  IROS}, \bibinfo{organization}{IEEE}. pp. \bibinfo{pages}{151--156}.
%Type = Inproceedings
\bibitem[{Patel et~al.(2017)}]{patel2017sensor}
\bibinfo{author}{Patel}, et~al., \bibinfo{year}{2017}.
\newblock \bibinfo{title}{Sensor modality fusion with cnns for ugv autonomous
  driving in indoor environments}, in: \bibinfo{booktitle}{IROS, IEEE}.
%Type = Inproceedings
\bibitem[{{Poria} et~al.(2016){Poria}, {Chaturvedi}, {Cambria} and
  {Hussain}}]{7837868}
\bibinfo{author}{{Poria}, S.}, \bibinfo{author}{{Chaturvedi}, I.},
  \bibinfo{author}{{Cambria}, E.}, \bibinfo{author}{{Hussain}, A.},
  \bibinfo{year}{2016}.
\newblock \bibinfo{title}{Convolutional mkl based multimodal emotion
  recognition and sentiment analysis}, in: \bibinfo{booktitle}{2016 IEEE 16th
  International Conference on Data Mining (ICDM)}, pp.
  \bibinfo{pages}{439--448}.
\newblock \DOIprefix\doi{10.1109/ICDM.2016.0055}.
%Type = Article
\bibitem[{Ramachandram et~al.(2017)}]{ramachandram2017deep}
\bibinfo{author}{Ramachandram}, et~al., \bibinfo{year}{2017}.
\newblock \bibinfo{title}{Deep multimodal learning: A survey on recent advances
  and trends}.
\newblock \bibinfo{journal}{IEEE Signal Processing Magazine}
  \bibinfo{volume}{34}, \bibinfo{pages}{96--108}.
%Type = Article
\bibitem[{Stein et~al.(2014)}]{Stein:NatureNeuro:2014}
\bibinfo{author}{Stein, B.E.}, et~al., \bibinfo{year}{2014}.
\newblock \bibinfo{title}{Development of multisensory integration from the
  perspective of the individual neuron}.
\newblock \bibinfo{journal}{Nature reviews. Neuroscience} \bibinfo{volume}{15},
  \bibinfo{pages}{520--535}.
%Type = Article
\bibitem[{Wei et~al.(2018)}]{wei2018LIDAR}
\bibinfo{author}{Wei}, et~al., \bibinfo{year}{2018}.
\newblock \bibinfo{title}{Lidar and camera detection fusion in a real-time
  industrial multi-sensor collision avoidance system}.
\newblock \bibinfo{journal}{Electronics} \bibinfo{volume}{7},
  \bibinfo{pages}{84}.
%Type = Inproceedings
\bibitem[{You et~al.(2017)}]{lattice}
\bibinfo{author}{You, S.}, et~al., \bibinfo{year}{2017}.
\newblock \bibinfo{title}{Deep lattice networks and partial monotonic
  functions}, in: \bibinfo{booktitle}{NIPS}.
\newblock \URLprefix \url{https://arxiv.org/abs/1709.06680}.
%Type = Article
\bibitem[{Yu et~al.(2019)Yu, Cuppini, Xu, Rowland and
  Stein}]{Yu:JNEUROSCI:2019}
\bibinfo{author}{Yu, L.}, \bibinfo{author}{Cuppini, C.}, \bibinfo{author}{Xu,
  J.}, \bibinfo{author}{Rowland, B.A.}, \bibinfo{author}{Stein, B.E.},
  \bibinfo{year}{2019}.
\newblock \bibinfo{title}{Cross-modal competition: The default computation for
  multisensory processing}.
\newblock \bibinfo{journal}{Journal of Neuroscience} \bibinfo{volume}{39},
  \bibinfo{pages}{1374--1385}.
\newblock \URLprefix \url{https://www.jneurosci.org/content/39/8/1374},
  \DOIprefix\doi{10.1523/JNEUROSCI.1806-18.2018},
  \href{http://arxiv.org/abs/https://www.jneurosci.org/content/39/8/1374.full.pdf}{\tt
  arXiv:https://www.jneurosci.org/content/39/8/1374.full.pdf}.
%Type = Inproceedings
\bibitem[{Zhang et~al.(2019)Zhang, Zhou, Sun, Wang, Shi and
  Loy}]{zhang2019robust}
\bibinfo{author}{Zhang, W.}, \bibinfo{author}{Zhou, H.}, \bibinfo{author}{Sun,
  S.}, \bibinfo{author}{Wang, Z.}, \bibinfo{author}{Shi, J.},
  \bibinfo{author}{Loy, C.C.}, \bibinfo{year}{2019}.
\newblock \bibinfo{title}{Robust multi-modality multi-object tracking}, in:
  \bibinfo{booktitle}{Proceedings of the IEEE International Conference on
  Computer Vision}, pp. \bibinfo{pages}{2365--2374}.

\end{thebibliography}


\bio{}
Myung Seok Shim received his Ph.D degree in Computer Engineering at Texas A\&M University, USA. Now, he works as an AI resident at Shell. His research interests are in deep learning for data fusion and interpretable neural network.
\endbio

\bio{}
Chenye Zhao received his M.S degree in Computer Engineering at Texas A\&M University, USA. Now, he is pursuing a Ph.D at University of Chicago, Illinois, USA.
\endbio

\bio{}
Yang Li received his M.S degree in Computer Engineering at Texas A\&M University, USA. Now, he works as a software development engineer at Amazon.
\endbio

\bio{}
Xuchong Zhang received his Ph.D degree in Electrical and Computer Engineering and he works as an assistant professor at Xi'an Jiaotong University, P.R. China. 
\endbio

\bio{}
Wenrui Zhang is a Ph.D student at University of California, Santa Barbara, USA. He studies spiking neural network architecture.
\endbio

\bio{}
Peng Li is a Professor of Electrical and Computer Engineering, University of California, Santa Barbara, USA. He received his  Ph.D in electrical and computer engineering from Carnegie Mellon University, Pittsburgh, PA. His research interests are in integrated circuits and systems,  learning algorithms and circuits for neuromorphic (a.k.a brain-inspired) computing, VLSI computer-aided design, machine learning and its hardware realization in VLSI, and computational brain modeling. 
\endbio

\end{document}